\documentclass[runningheads]{llncs}
\usepackage[T1]{fontenc}
\usepackage{graphicx,verbatim}
\usepackage{amsmath,amssymb}
\begin{document}
\title{A Cycle-Consistent Graph Surrogate for Full-Cycle Left Ventricular Myocardial Biomechanics}

\author{Siyu Mu\inst{1} \and
Wei Xuan Chan\inst{1} \and
Choon Hwai Yap\inst{1}}
\authorrunning{S.Mu et al.}
\institute{1. Department of Bioengineering, Imperial College London, UK \\siyu.mu21@imperial.ac.uk, c.yap@imperial.ac.uk}

\titlerunning{CGFENet}
\maketitle              % typeset the header of the contribution
\begin{abstract}
Image-based patient-specific simulation of left ventricular (LV) mechanics is valuable for understanding cardiac function and supporting clinical intervention planning, but conventional finite-element analysis (FEA) is computationally intensive. Current graph-based surrogates do not have full-cycle prediction capabilities, and physics-informed neural networks often struggle to converge on complex cardiac geometries. We present CardioGraphFENet (\textit{CGFENet}), a unified graph-based surrogate for rapid full-cycle estimation of LV myocardial biomechanics, supervised by a large FEA simulation dataset. The proposed model integrates (i) a global--local graph encoder to capture mesh features with weak-form-inspired global coupling, (ii) a gated recurrent unit-based temporal encoder conditioned on the target volume--time signal to model cycle-coherent dynamics, and (iii) a cycle-consistent bidirectional formulation for both loading and inverse unloading within a single framework. These strategies enable high fidelity with respect to traditional FEA ground truths and produce physiologically plausible pressure--volume loops that match FEA results when coupled with a lumped-parameter model. In particular, the cycle-consistency strategy enables a significant reduction in FEA supervision with only minimal loss in accuracy. Codes are available at \url{https://github.com/SiyuMU/CardioGraphFENet}.

% Authors must provide keywords and are not allowed to remove this Keyword section.

\end{abstract}

\section{Introduction}
Understanding cardiac mechanics is fundamental to characterising cardiovascular disease and forms the basis for building cardiac digital twins that enable personalised diagnosis and treatment planning \cite{zou2025digital,coorey2022health}. The traditional approach is image-based finite-element analysis (FEA), which couples image-derived patient anatomy with myocardial constitutive and contractile tension models to estimate deformation and stress across the cardiac cycle \cite{hoang2025finite,silva2014personalised}. However, because these simulations involve nonlinear material behaviour, large-deformation kinematics, and high-resolution meshes, runtimes are often prohibitively high \cite{guan2021mathematical,costa1996three,niederer2011simulating,lashgari2024patient}. The computational burden is further amplified by iterative inverse computations required to estimate the zero-pressure reference state of the heart, as well as by the need to simulate multiple cardiac cycles to eliminate initial-condition artefacts. Hence, patient-specific modelling remains difficult to deploy within clinically acceptable timescales \cite{gray2018patient,capelli2018patient}, motivating the need for fast deep learning (DL)-based FEA surrogates.

Several studies have developed such FEA surrogates to reduce computational cost. Early work used standard machine learning (ML) models (e.g., basic multilayer perceptrons (MLPs) and boosted trees) to predict quantities such as wall stress, deformation, and pressure–volume (P--V) relations, showing large speed-ups over FEA \cite{liang2018deep,dabiri2019prediction,maso2020deep,dabiri2020application}. However, these models often assume fixed mesh topology or require heavy geometric normalisation, and are typically limited to a single forward or inverse task. More recent efforts include physics-informed neural networks (PINNs) and graph neural networks (GNNs). PINNs enforce governing equations during training, but optimisation is still expensive and convergence can be challenging for complex cardiac PDEs. As such, they are currently limited to simplified geometries or reduced-order representations and are typically trained in a patient-specific manner rather than pre-trained \cite{buoso2021personalising,motiwale2024neural,mu2025imc}. Even with reduced-order models, per-case computation times remain substantial, particularly for finer meshes, which is not yet ideal. In contrast, GNNs trained on large simulation datasets allow real-time predictions without sacrificing detail through reduced-order modelling. They can handle unstructured FEA meshes with varying connectivity, and recent implementations have shown promising accuracy and speed \cite{dalton2023physics,dalton2022emulation,shi2025heartsimsage,mu2026heartunloadnet}. However, current GNN implementations do not model the full cardiac cycle, they do not include active contractile forces and are limited to passive loading during diastole or to estimation of the zero-pressure reference state \cite{mu2026heartunloadnet}. A unified, full-cycle surrogate applicable to diverse anatomies that can be adapted for inverse computing tasks within the same framework is needed.

In this work, we address this gap and develop such a DL surrogate framework for the left ventricle (LV). The model is generalisable across diverse ventricular geometries, and although it is trained with fixed myocardial stiffness and active tension generation, it can be extended through further training to accommodate different stiffnesses and active tension settings, as well as to the right ventricle (RV). The proposed framework, CardioGraphFENet (\textit{CGFENet}), utilises a GNN-based mesh encoder that captures anatomical features from image-derived unstructured LV geometries, and a temporal encoder that processes the target cavity volume and cardiac time. This design enables the model to infer pressure and full-field displacements from volume–time (V--t) inputs, and to handle both loading and unloading within a cycle-consistent architecture.

Our contributions are threefold: (1) we propose a cycle-consistent unified surrogate that jointly supports forward loading (predicting pressure and displacement from prescribed V--t inputs) and inverse unloading (recovering the zero-pressure reference state), (2) we enable mesh-agnostic full-cycle LV mechanics by predicting pressure and deformation over the entire diastolic--systolic loop on arbitrary LV meshes without registration or reduced-order representations, and (3) we incorporate lumped-parameter coupling to reconstruct physiologically plausible closed P--V loops consistent with the predicted mechanics.

\begin{figure}[t]
	\centering
	\includegraphics[width=\textwidth]{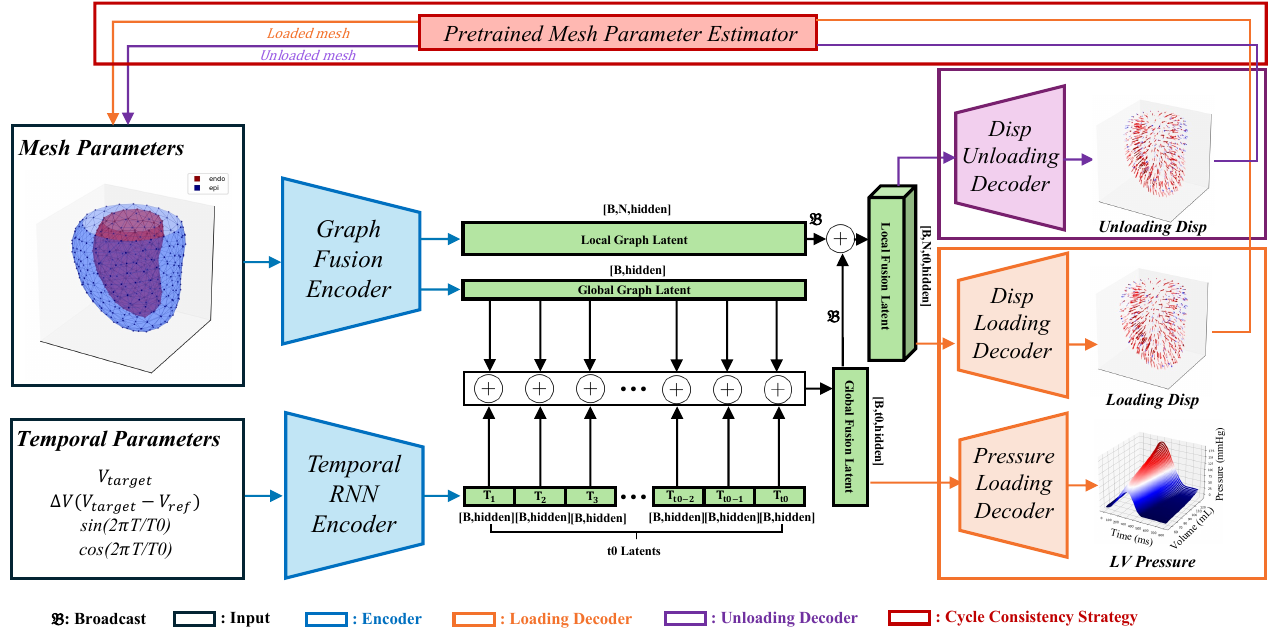}
    \caption{Overall architecture of \textit{CGFENet}. A dual-stream design encodes LV geometry (graph encoder) and the V--t signal (temporal encoder) into a shared latent space. The fused latent drives two task heads for forward loading (pressure and displacement) and inverse unloading (zero-pressure recovery), while a cycle-consistent constraint regularises the two mappings across the cardiac cycle. A pre-trained mesh-parameter estimator (frozen) provides the shared global features used in Fig.~\ref{FIG:Encoder}.}
	\label{FIG:Schematic}
\end{figure}

\section{Methods}
\subsection{Problem Domain}
In LV FEA, the loading state is described by cavity pressure $P_t$, cavity volume $V_t$, and cardiac time $t$. Given the unloaded reference configuration $\Omega_0$ and fixed constitutive and boundary conditions, specifying any two of $(P_t,V_t,t)$ determines the third and the displacement field $\mathbf{u}_t$. In this work, we choose $(V_t,t)$ as inputs and learn a dense mapping $(V_t,t)\mapsto (P_t,\mathbf{u}_t)$, which enables efficient generation of cycle-wide P--V--t trajectories from image-derived V--t curves or coupled to a lumped-parameter model. We also consider the inverse unloading task and recover the zero-pressure configuration $\Omega_0$ from an image-derived loaded geometry $\Omega_t^{\mathrm{load}}$ at the corresponding phase $t$. Thus, LV biomechanics involves two complementary operators:

\begin{equation}
\begin{aligned}
\mathcal{F}_{\mathrm{load}}&:\; (V_t,\, t,\, \Omega_0)\rightarrow \{P_t,\, \mathbf{u}_t\},\\
\mathcal{F}_{\mathrm{unload}}&:\; (\Omega_t^{\mathrm{load}},\, t)\rightarrow \Omega_0.
\end{aligned}
\end{equation}

% The aim of this work is to replace both high-cost FEA operators with a unified surrogate capable of performing forward loading and zero-pressure unloading on arbitrary LV meshes.

\subsection{\textit{CGFENet} Architecture}

\begin{figure}[t]
	\centering
	\includegraphics[width=\textwidth]{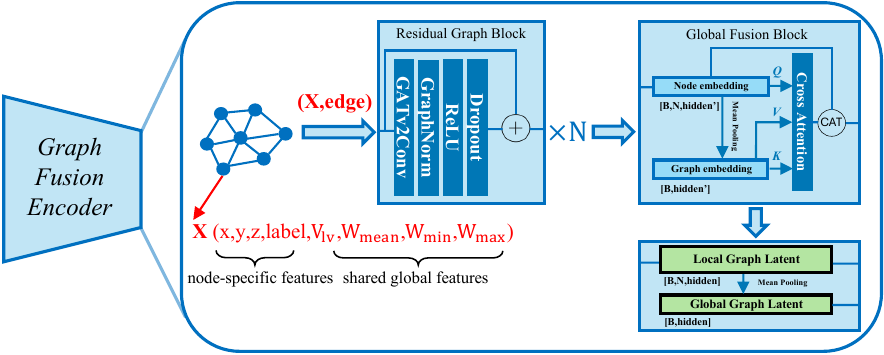}
	\caption{Graph Fusion Encoder. LV mesh is represented as $(\mathbf{X},\mathrm{edge})$ with node-wise features (coordinates and labels) and shared global descriptors $(V_{LV},W_{\mathrm{mean}},W_{\min},W_{\max})$, applies $N$ stacked residual GATv2 blocks, and injects FEA-inspired global coupling via a mean-pooled chamber token fused back to nodes through cross-attention, producing local and global graph latents.}
	\label{FIG:Encoder}
\end{figure}
Fig.~\ref{FIG:Schematic} illustrates \textit{CGFENet}, a unified dual-stream surrogate that approximates both forward loading and inverse unloading within a single cycle-consistent framework. \textit{CGFENet} takes (i) an unstructured LV mesh and (ii) a prescribed V--t signal as inputs, and outputs the cycle-wide cavity pressure and full-field displacement for loading, as well as the zero-pressure configuration for unloading.

\textbf{Graph Fusion Encoder:}
The Graph Fusion Encoder (Fig.~\ref{FIG:Encoder}) takes an LV mesh as input, represented as a graph $(\mathbf{X},\mathrm{edge})$ with node features $\mathbf{X}\in\mathbb{R}^{B\times N\times 8}$ for a batch of $B$ meshes with $N$ nodes. Each node feature concatenates local attributes $(x,y,z,\mathrm{label})$ and shared global descriptors $(V_{LV},W_{\mathrm{mean}},W_{\min},W_{\max})$ broadcast to all nodes, where $\mathrm{label}\in\{0,1,2\}$ denotes endocardium/epicardium/\\interior, $V_{LV}$ is the cavity volume, and $W_{\mathrm{mean}},W_{\min},W_{\max}$ are the mean/min/m-\\ax myocardial wall thickness.

The node embedding is initialised as $\mathbf{h}_i^{(0)}=\mathbf{X}_i$ (where $i\in\{1,\dots,N\}$) and updated by stacked GATv2 \cite{brody2021attentive} residual blocks. We use $M{=}4$ attention heads with a residual update:
\begin{equation}
\mathbf{h}_i^{(\ell+1)}
=
\mathbf{h}_i^{(\ell)}
+
\mathrm{Dropout}\!\Bigg(
\mathrm{ReLU}\!\Bigg(
\mathrm{GraphNorm}\!\Bigg(
\big\Vert_{m=1}^{M}
\sum_{j\in\mathcal{N}(i)}
\alpha_{ij}^{(m)}\, W^{(m)} \mathbf{h}_j^{(\ell)}
\Bigg)
\Bigg)
\Bigg),
\end{equation}
where $\alpha_{ij}^{(m)}$ is the GATv2 attention weight (softmax-normalised over $j\in\mathcal{N}(i)$).

To enable both local message passing and global awareness, in which element-wise weak-form terms are aggregated into a global equilibrium, we inject explicit global coupling into the graph encoder. Inspired by GraphGPS-style \cite{rampavsek2022recipe} designs, we summarise the chamber with a lightweight global token via mean pooling and fuse it back to all nodes using global-to-local attention. This global pathway helps the encoder capture FEA-like global consistency (e.g., P--V coupling). The encoder outputs node-wise local graph latent $\mathbf{Z}_{\mathrm{local}}$ for displacement decoding and a pooled global graph latent $\mathbf{Z}_{\mathrm{global}}$ for pressure prediction and temporal fusion.

\textbf{Temporal RNN Encoder:}
The temporal encoder models cycle-coherent dynamics under a prescribed V--t signal. For each time sample $t_k$ in a cardiac cycle, we form a compact input:
\begin{equation}
\mathbf{T}_k^{\mathrm{in}}=\Big(V_{\mathrm{target}},\,\Delta V,\,\sin(2\pi t_k/t_0),\,\cos(2\pi t_k/t_0)\Big),
\quad \Delta V=V_{\mathrm{target}}-V_{\mathrm{ref}},
\end{equation}
where $V_{\mathrm{target}}$ is the prescribed target LV cavity volume, $\Delta V$ is the volume offset from the mesh reference cavity volume $V_{\mathrm{ref}}$, $t_k$ is the $k$-th time sample in the cardiac cycle, and $t_0$ is the cycle length used for temporal normalisation.

An MLP first embeds $\mathbf{T}_k^{\mathrm{in}}$ into a time-conditioned feature, and a gated recurrent unit (GRU) propagates it over the cycle to produce a temporal latent sequence $\{\mathbf{T}_k\}_{k=1}^{t_0}$. This latent captures smooth, history-dependent dynamics and is used for global fusion and pressure/displacement prediction. During training, temporal modelling is performed at a fixed target volume across time, allowing the network to learn cycle-dependent dynamics independently of instantaneous volume variation.

\textbf{Cycle Consistent Strategy:}
\textit{CGFENet} fuses graph and temporal latents $(\mathbf{Z}_{\mathrm{local}}, \mathbf{Z}_{\mathrm{global}},\{\mathbf{T}_k\}_{k=1}^{t_0})$ to form a global representation $\mathbf{L}_{\mathrm{global}}$ for pressure prediction and a spatio-temporal representation $\mathbf{L}_{\mathrm{local}}$ for nodal displacement prediction:
\begin{equation}
\begin{aligned}
\mathbf{L}_{\mathrm{global}} &= \left\{ W_{\mathrm{global}}\!\left(\mathbf{Z}_{\mathrm{global}} + \mathbf{T}_k\right) \right\}_{k=1}^{t_0} &&\in \mathbb{R}^{B \times t_0 \times hidden}\\
\mathbf{L}_{\mathrm{local}} &= W_{\mathrm{local}}\!\left(\mathfrak{B}(\mathbf{Z}_{\mathrm{local}})+\mathfrak{B}(\mathbf{L}_{\mathrm{global}})\right) &&\in\mathbb{R}^{B\times N\times t_0\times hidden}
\end{aligned}
\end{equation}

To couple loading and unloading in a single network (between a state in the cardiac cycle and the unloaded reference), we enforce cycle consistency during training. The key idea is that, at any phase $t$, the loading ($\Omega_0 \mapsto \Omega_t^{\mathrm{load}}$) and unloading ($\Omega_t^{\mathrm{load}} \mapsto \Omega_0$) branches should behave as approximate inverses. Therefore, we follow a \emph{loading$\rightarrow$unloading$\rightarrow$re-loading} cycle during training. Starting from the zero-pressure mesh input $\mathbf{X}_{\mathrm{unload}}^{\ast}$, the loading branch predicts the loaded mesh $\mathbf{X}_{\mathrm{load}}^{(1)}$ and pressure $P_t^{(1)}$. We then feed $\mathbf{X}_{\mathrm{load}}^{(1)}$ into the unloading branch to obtain $\mathbf{X}_{\mathrm{unload}}^{(1)}$, and re-apply the loading branch on $\mathbf{X}_{\mathrm{unload}}^{(1)}$ to reconstruct $\mathbf{X}_{\mathrm{load}}^{(2)}$ and $P_t^{(2)}$. Cycle consistency is enforced by penalising the mismatch between $\mathbf{X}_{\mathrm{load}}^{(2)}$ and $\mathbf{X}_{\mathrm{load}}^{(1)}$, and between $\mathbf{X}_{\mathrm{unload}}^{(1)}$ and $\mathbf{X}_{\mathrm{unload}}^{\ast}$, encouraging the two branches to learn mutually inverse mappings while reinforcing the capability of the shared encoder. Alongside this cycle constraint, we apply paired FE supervision by matching the first-pass predictions $(\mathbf{X}_{\mathrm{load}}^{(1)}, P_t^{(1)})$ and $\mathbf{X}_{\mathrm{unload}}^{(1)}$ to their FE references $(\mathbf{X}_{\mathrm{load}}^{\ast}, P_t^{\ast})$ and $\mathbf{X}_{\mathrm{unload}}^{\ast}$. From our experiments below, the cycle term indeed acts as an additional regulariser, improving invertibility and significantly reducing the amount of paired supervision required to reach a given accuracy.

\textbf{Inference Procedure:}
At inference, we reconstruct an LV mesh from imaging (typically at end-diastole (ED)), predict its zero-pressure configuration via the unloading branch, and then query the loading branch over a dense set of (V,t) to obtain a patient-specific P--V--t lookup table. A lumped-parameter model~\cite{cairelli2024role} then closes the P--V loop by querying this table, enabling efficient circulatory updates without iterative FEA (or surrogates) solving.

\textbf{Data Generation:}
We generated an LV mechanics FEA dataset by sampling ED LV geometries from an image-derived Principal Component Analysis (PCA) statistical shape model trained on 1,991 asymptomatic subjects~\cite{medrano2014left}, with coefficients truncated to $[-2,2]$. Sampled ED surfaces were converted into patient-specific tetrahedral FE meshes using Gmsh~\cite{geuzaine2009gmsh}, the resulting ED cavity volumes and myocardial wall-thickness statistics broadly fell within reported physiological ranges~\cite{hudsmith2005normal,walpot2019left}. For each mesh, we performed inverse-FE unloading to obtain the zero-pressure reference configuration, then ran forward FEA from this unloaded state over wide range of cavity volumes (40--160\,mL, 50 samples) and all time points within the cardiac cycle (0-800\,ms, 800 samples) to compute chamber pressure and nodal displacement fields. All simulations used the same constitutive model (Fung-type \cite{guccione1991passive}) and adult parameter setting as in~\cite{mu2025imc}. Overall, we generated 67 anatomically distinct meshes and 1,814,146 supervised $P$–$V$–$t$ states, rare cases of non-convergence were discarded. We split the data at the mesh (subject) level (approximately 4:1 train/test), keeping all states from the same mesh within a single subset.

\textbf{Implementation:}
All models were implemented in PyTorch and trained on an NVIDIA RTX~3090 (24\,GB). We used hidden width 128 with a GATv2 backbone (4 heads) and dropout 0.1. Optimisation was performed with AdamW (learning rate $1\times10^{-3}$, weight decay $1\times10^{-3}$). The total loss used fixed weights $\lambda_{p}=1$ and $\lambda_{d}=10^{4}$ for pressure and displacement, respectively.

\section{Results}
Table~\ref{tab:ablation_all} first benchmarks \textit{CGFENet} against two popular graph surrogates, GraphUNet (B1)~\cite{gao2019graph} and MeshGraphNet (B2)~\cite{pfaff2020learning}. 
\textit{CGFENet} (A0) achieves the best performance in both deformation fidelity (lower RMSE/HD and higher vertex accuracy) and pressure prediction (higher $R^2$ and lower RMSE) over both baselines. Ablations further isolate the contribution of each module. Replacing the GRU with a non-recurrent temporal encoder (A1) keeps displacement comparable but substantially degrades pressure prediction (RMSE $1.70\!\rightarrow\!2.43$\,mmHg, $\sim$1.4$\times$) and produces less smooth cycle-wide waveforms, highlighting the importance of recurrent temporal modelling. Removing global fusion (A2) consistently worsens both loading and unloading, confirming the need for explicit chamber-level coupling beyond local message passing. Removing cycle consistency (A3) most strongly impacts unloading (largest HD), indicating that the bidirectional constraint is critical for learning a stable near-inverse mapping.

To assess supervision efficiency (Table~\ref{tab:weaksup}), we vary the supervision ratio (SR), defined as the fraction of training meshes that have FE labels (all meshes are used during training). Labelled meshes contribute supervised terms and the cycle-consistency term, whereas unlabelled meshes contribute only via the cycle-consistency term. With cycle consistency, loading vertex accuracy ($<1$\,mm) remains high for SR $=70\%$--$30\%$ and degrades gracefully at SR $=10\%$. Without the cycle constraint, performance collapses under limited labels, dropping sharply at SR $=30\%$ and further at SR $=10\%$. These results indicate that cycle consistency provides an effective self-regularisation signal that substantially reduces reliance on paired FE supervision.

\begin{table}[t]
\caption{Baseline and ablation results of \textit{CGFENet} on displacement and pressure prediction. Acc.\ denotes vertex accuracy, which is the fraction of vertices with Euclidean displacement error below 1 mm or 2 mm.}
\label{tab:ablation_all}
\centering
\scriptsize
\begin{tabular}{|l|p{3.6cm}|c|c|c|c|}
\hline

\multicolumn{6}{|c|}{\textbf{Forward (Loading) -- Displacement}}\\
\hline
Model & Description & RMSE (mm)$\downarrow$ & Acc.\,$<1$mm$\uparrow$ & Acc.\,$<2$mm$\uparrow$ & HD (mm)$\downarrow$\\
\hline
A0 & \textbf{\textit{CGFENet}}
& $0.39 \pm 0.01$
& $\mathbf{0.90 \pm 0.01}$
& $\mathbf{0.99 \pm 0.02}$
& $1.89 \pm 0.15$\\
\hline
A1 & w/o GRUs
& $\mathbf{0.34 \pm 0.01}$
& $0.89 \pm 0.01$
& $\mathbf{0.99 \pm 0.02}$
& $3.13 \pm 0.19$\\
\hline
A2 & w/o global fusion module
& $0.43 \pm 0.01$
& $0.84 \pm 0.02$
& $0.98 \pm 0.02$
& $3.31 \pm 0.22$\\
\hline
A3 & w/o cycle-consistency
& $0.38 \pm 0.01$
& $0.88 \pm 0.01$
& $0.98 \pm 0.01$
& $\mathbf{1.58 \pm 0.12}$\\
\hline
B1 & GraphUNet 
& $0.42 \pm 0.02$
& $0.85 \pm 0.01$
& $0.87 \pm 0.02$
& $3.30 \pm 0.11$\\
\hline
B2 & MeshGraphNet
& $0.44 \pm 0.02$
& $0.81 \pm 0.03$
& $0.83 \pm 0.03$
& $4.88 \pm 0.24$\\
\hline

\multicolumn{6}{|c|}{\textbf{Inverse (Unloading) -- Displacement}}\\
\hline
Model & Description & RMSE (mm)$\downarrow$ & Acc.\,$<1$mm$\uparrow$ & Acc.\,$<2$mm$\uparrow$ & HD (mm)$\downarrow$\\
\hline
A0 & \textbf{\textit{CGFENet}}
& $0.30 \pm 0.02$
& $\mathbf{0.93 \pm 0.01}$
& $\mathbf{0.99 \pm 0.03}$
& $\mathbf{1.38 \pm 0.12}$\\
\hline
A1 & w/o GRUs
& $\mathbf{0.20 \pm 0.02}$
& $0.92 \pm 0.01$
& $\mathbf{0.99 \pm 0.03}$
& $1.52 \pm 0.11$\\
\hline
A2 & w/o global fusion module
& $0.37 \pm 0.02$
& $0.86 \pm 0.02$
& $0.98 \pm 0.06$
& $1.75 \pm 0.12$\\
\hline
A3 & w/o cycle-consistency 
& $0.41 \pm 0.03$
& $0.89 \pm 0.02$
& $0.97 \pm 0.08$
& $3.21 \pm 0.12$\\
\hline

\multicolumn{6}{|c|}{\textbf{Forward (Loading) -- Pressure}}\\
\hline
Model & Description 
& \multicolumn{2}{c|}{$R^2\uparrow$}
& \multicolumn{2}{c|}{RMSE (mmHg)$\downarrow$}\\
\hline
A0 & \textbf{\textit{CGFENet}}
& \multicolumn{2}{c|}{$\mathbf{0.98 \pm 0.01}$}
& \multicolumn{2}{c|}{$\mathbf{1.70 \pm 0.17}$}\\
\hline
A1 & w/o GRUs
& \multicolumn{2}{c|}{$\mathbf{0.98 \pm 0.01}$}
& \multicolumn{2}{c|}{$2.43 \pm 0.18$}\\
\hline
A2 & w/o global fusion module
& \multicolumn{2}{c|}{$\mathbf{0.98 \pm 0.01}$}
& \multicolumn{2}{c|}{$2.12 \pm 0.18$}\\
\hline
A3 & w/o cycle-consistency
& \multicolumn{2}{c|}{$\mathbf{0.98 \pm 0.01}$}
& \multicolumn{2}{c|}{$2.03 \pm 0.21$}\\
\hline
B1 & GraphUNet 
& \multicolumn{2}{c|}{$0.95 \pm 0.01$}
& \multicolumn{2}{c|}{$5.83 \pm 0.22$}\\
\hline
B2 & MeshGraphNet 
& \multicolumn{2}{c|}{$\mathbf{0.98 \pm 0.01}$}
& \multicolumn{2}{c|}{$3.63 \pm 0.10$}\\
\hline

\end{tabular}
\end{table}

\begin{table}[t]
\caption{Vertex accuracy (<1mm) for loading prediction under reduced supervision}
\label{tab:weaksup}
\centering
\scriptsize
\begin{tabular}{|c|c|c|c|c|}
\hline
\textbf{SR} & \textbf{70\%} & \textbf{50\%} & \textbf{30\%} & \textbf{10\%} \\
\hline
\textbf{with cycle} & $\mathbf{0.89 \pm 0.01}$ & $0.88 \pm 0.01$ & $0.88 \pm 0.01$ & $0.70 \pm 0.02$ \\
\hline
\textbf{w/o cycle} & $0.80 \pm 0.02$ & $0.86 \pm 0.01$ & $0.50 \pm 0.03$ & $0.25 \pm 0.03$ \\
\hline
\end{tabular}
\end{table}

\begin{figure}[t]
	\centering
	\includegraphics[width=\textwidth]{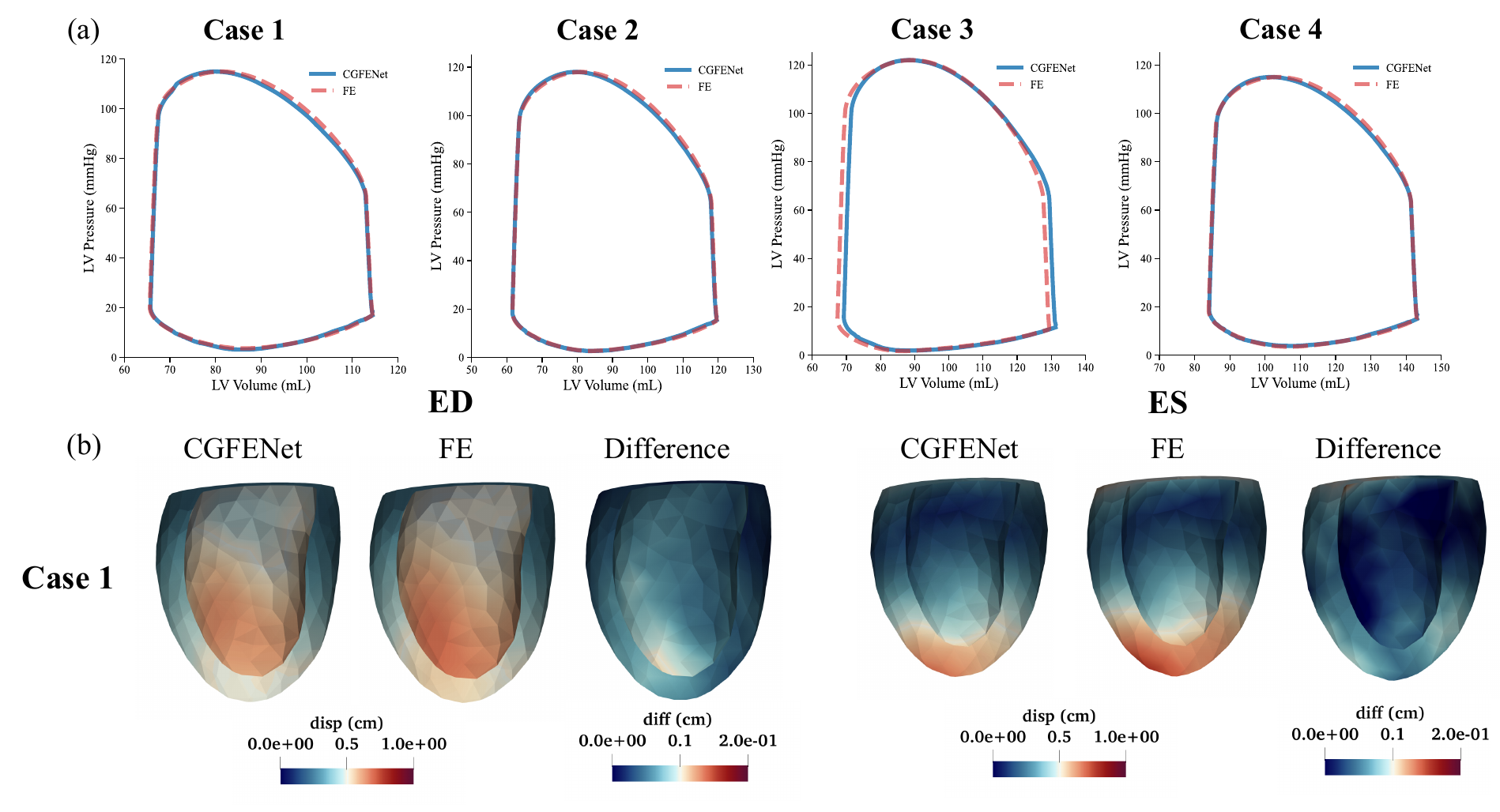}
    \caption{Qualitative full-cycle and geometry evaluation on representative cases. (a)  lumped-parameter-coupled P--V loops for four cases, comparing \textit{CGFENet} predictions with FE references. (b) Surface visualisations at ED and end-systole (ES) for one case: \textit{CGFENet} and FE show displacement magnitude relative to the zero-pressure configuration, while Difference reports the vertex-wise discrepancy (\textit{CGFENet} minus FE).}
	\label{FIG:vis}
\end{figure}

We assessed closed-loop P--V fidelity by coupling \textit{CGFENet} with a lumped-parameter model and iteratively querying the surrogate P--V--t table until the full-cycle trajectory converged to a stable closed loop. Fig.~\ref{FIG:vis} (a) shows close agreement with FEA across both systolic and diastolic branches for four representative cases, with slightly larger deviations in Case~3 around transition regions. Fig.~\ref{FIG:vis} (b) further visualises ED and ES deformation magnitudes relative to the unloaded reference and the residual (prediction minus FEA), indicating that \textit{CGFENet} preserves the expected spatially non-uniform motion with dominant apical deformation and generally small, localised errors.

\textit{CGFENet} generates a full case-level P--V--t map in $44.3\pm11.2$\,s, yielding an $\sim74\times$ speed-up over traditional FEA ($3256.8\pm1599.2$\,s).

\section{Discussion and Conclusion}
We presented \textit{CGFENet}, a unified surrogate for full-cycle LV mechanics that performs both forward loading and inverse unloading on variable tetrahedral meshes. Experiments show that \textit{CGFENet} improves over representative graph baselines in both deformation and pressure prediction. The gains come from combining attention-based graph encoding, global coupling of local connections, temporal modelling, and cycle-consistent training, where cycle consistency particularly strengthens the inverse mapping and improves label efficiency. Windkessel-coupled evaluation further confirms that the predicted pressures yield physiologically plausible P--V loops and consistent deformation patterns across cases.

Limitations mainly stem from the current dataset rather than the model design. In this study, simulations were generated with fixed stiffness and active tension settings, so inter-subject variability in these parameters is not yet explicitly represented. We are currently extending the dataset to include them, and we expect \textit{CGFENetv2} to accommodate these additional inputs with minor modifications.

As a downstream component for image-driven cardiac simulation in computer assisted intervention, the surrogate starts from an image reconstructed 3D LV mesh and efficiently predicts full-cycle mechanics and pressure, avoiding repeated non-linear FEA solves. This makes FEA-style biomechanical outputs, including full-field displacement, patient-specific pressure trajectories, and derived functional indices, available for time-critical workflows such as procedural planning, rapid what-if scenario analysis, and decision support, where conventional FE runtimes are prohibitive.

\bibliographystyle{splncs04}
\bibliography{mybibliography}

\end{document}